\documentclass[sigconf, nonacm = true]{acmart}
\usepackage{bbding}
\usepackage{multirow}
\usepackage[normalem]{ulem}
\useunder{\uline}{\ul}{}
\usepackage{graphicx}
\graphicspath{ {./images/} }
\usepackage{caption}
\usepackage{subcaption}
\usepackage{enumitem}
\usepackage{tabularx}

\setcopyright{acmcopyright}
\copyrightyear{2022}
\acmYear{2022}
\acmDOI{XXXXXXX.XXXXXXX}

\acmConference[Conference acronym 'XX]{Make sure to enter the correct
  conference title from your rights confirmation emai}{June 03--05,
  2018}{Woodstock, NY}
\acmPrice{15.00}
\acmISBN{978-1-4503-XXXX-X/18/06}

\begin{document}

\title{TransFA: Transformer-based Representation for \\ Face Attribute Evaluation}
\author{Decheng Liu, Weijie He, Chunlei Peng, Nannan Wang, Jie Li, Xinbo Gao}
\begin{abstract}
  Face attribute evaluation plays an important role in video surveillance and face analysis. Although methods based on convolution neural networks have made great progress, they inevitably only deal with one local neighborhood with convolutions at a time. Besides, existing methods mostly regard face attribute evaluation as the individual multi-label classification task, ignoring the inherent relationship between semantic attributes and face identity information. In this paper, we propose a novel \textbf{trans}former-based representation for \textbf{f}ace \textbf{a}ttribute evaluation method (\textbf{TransFA}), which could effectively enhance the attribute discriminative representation learning in the context of attention mechanism. 
  The multiple branches transformer is employed to explore the inter-correlation between different attributes in similar semantic regions for attribute feature learning. 
  Specially, the hierarchical identity-constraint attribute loss is designed to train the end-to-end architecture, which could further integrate face identity discriminative information to boost performance.
  Experimental results on multiple face attribute benchmarks demonstrate that the proposed TransFA achieves superior performances compared with state-of-the-art methods. 
   \emph{The code and models will be publicly available when the paper is accepted.}

\end{abstract}



\maketitle

\section{Introduction}

Face image analysis plays an important role in biometric security and computer vision filed. 
Here face attributes are regarded as the key semantic information, and these also could be applied in multiple real-world scenarios (e.g. surveillance, image retrieval and face manipulation) \cite{han2017heterogeneous,liu2015deep}. The core challenge of face attribute evaluation is to extract suitable representation which would effectively bridge visual words and image pixels in two different domains. Although great process has been achieved benefited by deep convolutional network \cite{mao2020deep,han2017heterogeneous,cao2018partially,hand2017attributes}, there still exist many challenges in real-world face attribute evaluation. Due to face images captured always under uncooperative scenarios, complex background, large posture change and illumination variations would unavoidably affect the face attribute estimation performance. Thus, face attribute evaluation is still a challenging problem. 

\begin{figure}[t]
    \centering
    \includegraphics[width=0.45\textwidth]{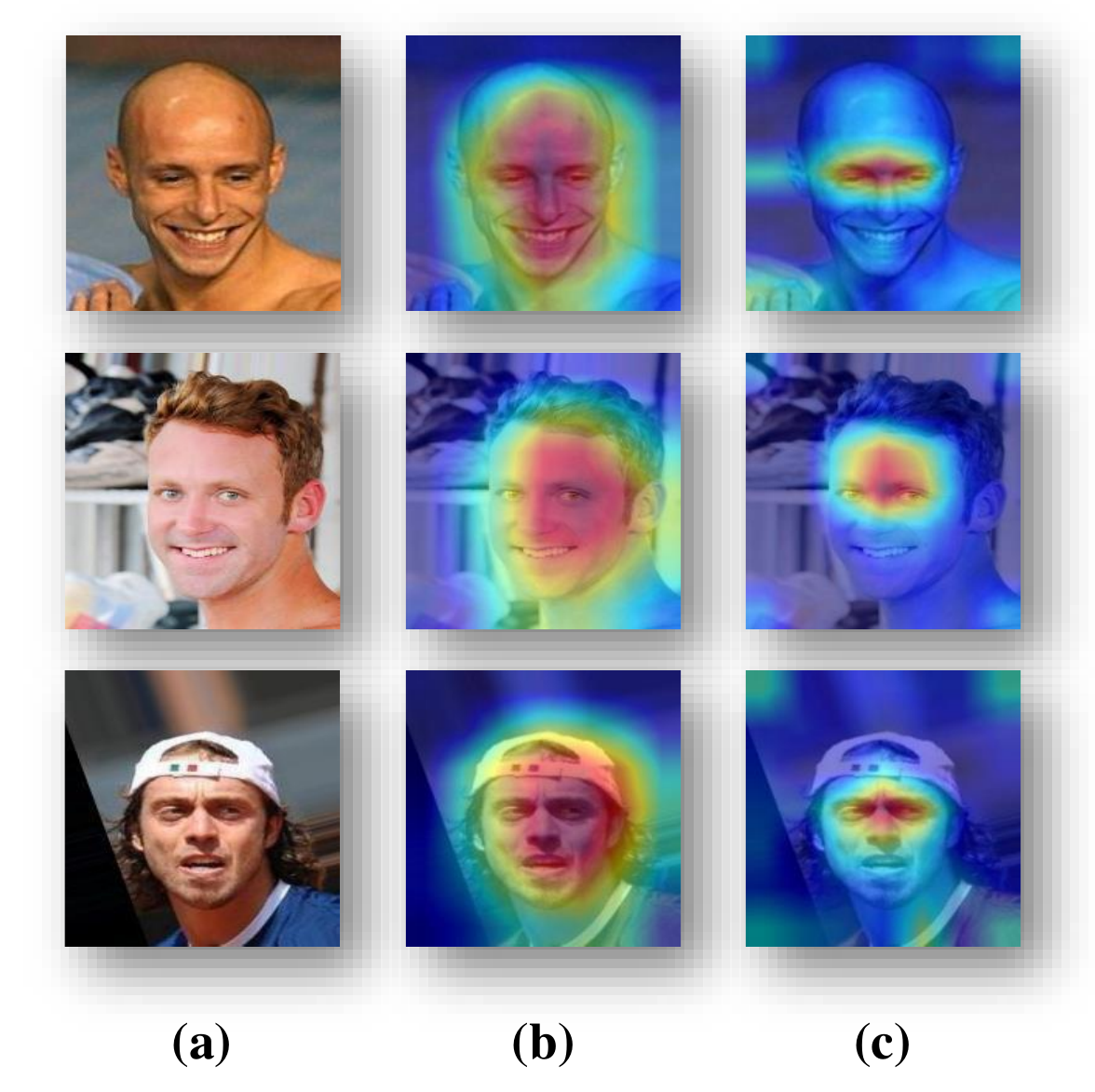}
    \caption{Grad-CAM visualization of different attention maps using different networks: (a) The original images, (b) CNN-based method, (c) Transformer-based method. Three facial images are evaluated as narrow eyes, bushy eyebrows and arched eyebrows from top to bottom respectively.}
    \label{fig:CNN&Trans}
\end{figure}

Existing face attribute evaluation methods could be classified into two categories: hand-crafted feature-based method and deep convolutional neural network (CNN) -based method. Early studies firstly extract hand-crafted feature, and then utilize designed classifier to estimate face attribute \cite{bourdev2011describing,luo2013deep,hwang2011sharing}. These approaches only perform well under constrained conditions differ from real application scenarios. With the development of deep learning in image analysis, especially CNN-based representation learning \cite{he2016deep,vggnet,inception,densent,senet}, researches focus on designing suitable discriminative features with CNN models \cite{nguyen2018large,li2015two,liu2015deep}. Besides, \cite{han2017heterogeneous} firstly explored the inter-correlations of face attributes, and the proposed joint attribute evaluation framework inspired other studies \cite{HuiDing2018ADC,mahbub2018segment,nguyen2018large} to construct face attribute shared feature learning module. It is noted that convolutions mainly focus on small local region due to receptive fields \cite{TransReID}. These downsampling operators of CNN also lose detailed discriminative information. Thus, there exists difficulties to explore effective attribute shared feature only with convolution operators. 
As shown in Figure \ref{fig:CNN&Trans}, the visualization of attention maps with CNN-based method show this kind of network makes it hard to focus on robust discriminative semantic parts in semantic regions. 
It is noted that we utilize the ResNet-based baseline to represent CNN-based method, and the proposed TransFA to represent Transformer-based method. More details are shown in Section \ref{Comparison with State-of-Art Methods}.

Recently, self-attention mechanism and transformer model \cite{TransReID} have drawn more and more attention on both natural language processing and computer vision tasks. Vision Transformer \cite{ViT,naseer2021intriguing} has shown the transformer-based model indeed contains the capacity as the backbone network instead of former pure CNN model in image synthesis and classification tasks \cite{he2016deep,szegedy2017inception}. 
The reason why the transformer-based model is more applicable than CNN-based model are shown as follow: the transformer architecture usually utilizes visual tokens to represents semantic concepts, and then model the relationship between them. This specific framework would be more effectively learn the long range dependencies in face semantic parts, which naturally helps to model the strong attribute correlations \cite{han2017heterogeneous}. As shown in Figure 1, the attention map visualization comparison illuminates that the transformer-based representation could easily observe the strong discriminative semantic parts in according attribute regions. Thus, it inspires us to design transformer model in face attribute evaluation in this paper. 

This paper proposes a novel transformer-based representation for face attribute evaluation (TransFA), which could learn the robust discriminative attribute feature representations. Firstly, in order to further model the strong inter-correlations between different attributes, we introduce shifted windows \cite{Swin} to design the shared attribute feature learning module, which could help model the connection between different semantic region even long-range correlations. Secondly, different from the most prior work designing attribute categories depending on prior knowledge (like data type, semantic meaning and subjectivity) \cite{han2017heterogeneous,mao2020deep}, the proposed method utilizes the semi-automatic attribute grouping strategy: visualizing attention maps of different attribute with Grad-CAM \cite{Grad-cam}, and then attributes with similar semantic regions are naturally categorized into the same group. Then, several local semantic region branches are designed with the multi-tasking learning in the attention-specific feature learning module. To further exploits the relationship between face attribute and identity information, we propose the global identity correlation loss in identity correlation branch, where global attribute feature is captured by fusing different local features. \emph{The main reason of superior performance is that we effectively introduce face attribute-specific inductive bias (e.g. attribute inter-correlations and face identity-related relationship) in designing transformer-based representation to enhance face attribute evaluation performance.}

To the best of our knowledge, it is  \emph{the first exploration} to introduce transformer model in the field of face attribute evaluation. The main contributions of our paper are summarized as follows:
\begin{enumerate}[itemsep= 20 pt, topsep = 5 pt]
    \setlength{\itemsep}{0pt}
    \item We firstly employ a transformer-based representation for face attribute evaluation, which could effectively integrate spatial inter-correlation between different attributes in similar semantic regions to enhance performance.
    \item The hierarchical identity-constraint attribute loss is designed by the inherent relationship between face identity and semantic attribute, which can make feature representations contain robust discriminative information.
    \item Experimental results illustrate the superior performance of the proposed TransFA compared with the state-of-the-art attribute evaluation methods.
     \emph{The code will be publicly available when the paper is accepted}.
\end{enumerate}

We organized the rest of this paper as follows. Section 1 gives a brief introduction of proposed method, and section 2 shows some representative face attribute evaluation algorithms. In Section 3, we present the novel transformer-based representation for face attribute evaluation. Section 4 shows the experimental results and analysis. The conclusion is drawn in Section 5. 

\section{RELATED WORK}
\subsection{Face Attribute Evaluation}
Here we review representative face attribute evaluation methods in the aforementioned two categories: hand-crafted feature-based methods and CNN feature-based methods.
Investigations on face attribute evaluation can be traced back to the
1990s \cite{cottrell1990empath}. Early approaches usually evaluate facial attributes with classic classifiers given the hand-crafted features, such as decision trees \cite{luo2013deep}, support vector machines \cite{cortes1995support, bourdev2011describing, kumar2009attribute} and k-nearest neighbor classifier \cite{huang2016learning, huang2019deep}. For instance, Kumar et al. \cite{kumar2009attribute} train multiple SVMs to predict multiple attributes with each SVM evaluating one face attribute. Luo et al. \cite{luo2013deep} investigate a sum-product decision tree network to locate the face attribute regions and evaluate face attributes at the same time. 

Recently with the development of deep learning techniques, 
more and more researchers focus on face attribute analysis field \cite{kumar2008facetracer, lei2011photo, zhang2014panda, liu2015deep, hand2017attributes, he2017adaptively, mahbub2018segment, mao2020deep, han2017heterogeneous}. 
Liu et al. \cite{liu2015deep} proposed a cascade deep neural network consisting of LNet and ANet for face attribute evaluation, where LNet aims at attribute localization and ANet aims at feature extraction. It is the first exploration in large-scale face attribute evaluation task simultaneously.
Hand et al. \cite{hand2017attributes} proposed a multi-task deep CNN model combing with an auxiliary network, which exploits attribute relationships to enhance performance. 
He et al. \cite{he2017adaptively} proposed an adaptive weighted multi-task deep network applying an adaptive and dynamical weighting strategy for face attribute evaluation. 
By considering attribute correlation and heterogeneity, Han et al. \cite{han2017heterogeneous} designed a Deep Multi-Task Learning Network (DMTL) to jointly evaluate different face attributes. 
Mao et al. \cite{mao2020deep} propose a deep multi-task multi-label CNN model, which could jointly optimize facial landmark detection task and facial attribute evaluation task to boost performance. However, researches on modeling the relationship between attribute and identity are limited, which may help networks better learn facial attribute discriminative features. 

In related face recognition task, identity could be regarded as the extra high-level semantic attribute for face analysis. And researches \cite{ACCVOuyang,ijcai18,TNNLS20,IF2017,ICCV2017} find that the extra attribute could help enhance representation discriminability.
Ouyang et al.\cite{ACCVOuyang} utilized the feature fusion strategy to directly fuse attribute and identity features for higher recognition accuracy.
Mittal et al.\cite{IF2017} designed the attribute feedback using reliable attribute to filter ranked list in the pre-processing stage.
Liu et al.\cite{ijcai18,TNNLS20} explored directly integrating attribute discriminative information with an end-to-end structure for recognition task.
Furthermore, Liu et al.\cite{LiuPR21} utilized the attribute clue to augment face sketches for increasing recognition accuracy.
Thus, there indeed exists the strong correlation between semantic attributes and face identity.
Inspired by the inherent relationship, we propose the hierarchical identity-constraint attribute loss to make representations more discriminative for face attribute analysis, which could effectively contain both high-level identity and semantic information.  

\subsection{Vision Transformer}
Transformer model is firstly proposed \cite{vaswani2017attention} in the natural language processing field  for machine translation tasks. 
Motivated by the success of the self-attention mechanism, numerous algorithms based on attention mechanism have been proposed in computer vision field recently. 
Alexey et al.\cite{ViT} firstly proposed the pure vision transformer for image classification task, which process the input images as the sequences of pixel patches.
Liu et al. \cite{Swin} investigated Swin Transformer containing shifted window and window-based self-attention mechanism, which could be served as a general backbone architecture. 
The pure transformer model has been drawn more and more attention in downstream vision tasks, including image classification, re-identification, object detection, tracking and image restoration.  \cite{han2020survey, TransReID, chen2021transformer, song2021vidt, liang2021swinir}.
He et al.\cite{TransReID} designed a transformer-based model for object re-identification, which could enhance the robust feature learning.
Liang et al.\cite{liang2021swinir} utilized Swin Transformer to construct feature extraction module, which could achieve superior performance in image restoration task. 
Though transformer-based methods have show impressive performance on several vision tasks, limited works have been conducted in face attribute analysis currently.
Consequently, the fine designed attribute-specific inductive bias are introduced in the proposed algorithm, and it could effectively model the inter-correlation between attribute to boost performance.

\section{PROPOSED APPROACH} \label{PROPOSED APPROACH}
\subsection{Motivation}

\begin{figure}[t]
    \centering
    \includegraphics[width=0.45\textwidth]{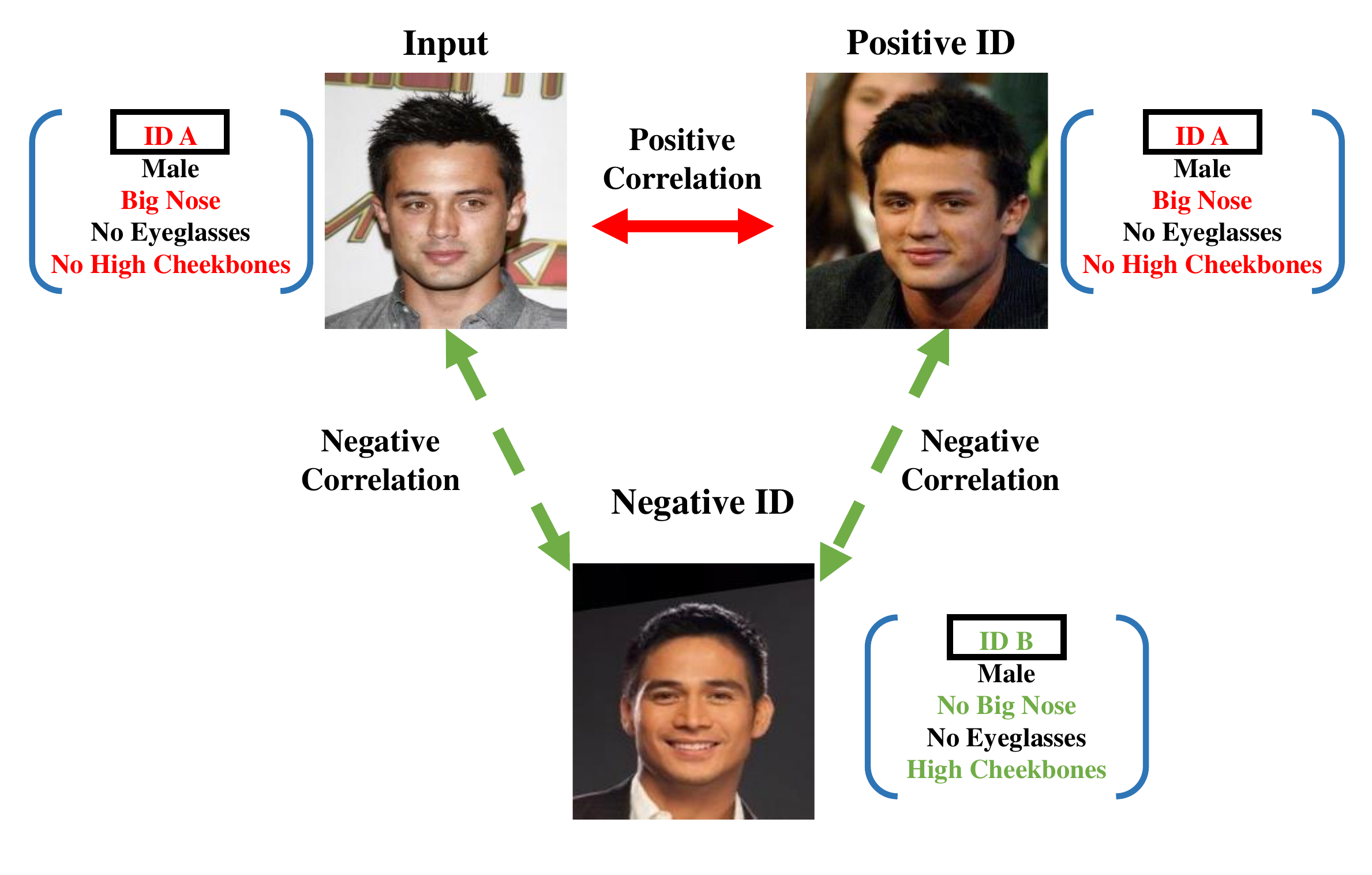}
    \caption{Facial images with the same identity usually have a positive attribute correlation, while images with different identities usually have a negative attribute correlation. }
    \label{fig:motivation}
\end{figure}

Motivated by the inherent relationship between semantic attribute and face identity, we propose the novel identity-correlation attribute loss to model the correlation. Here we give an example as shown in Figure \ref{fig:motivation}. For the input face, positive attribute correlations are marked with red color and negative correlations are marked with green color. For the convenience of observations, the identity information is regraded as the extra semantic information here. It is easy to find there indeed exists the inherent relationship between attributes and identity: 
 \emph{face images with the same identity always contain high-similarity attributes, and instead face images with different identities usually contain low-similarity attributes.} Thus, this inspires the high-level identity information could help enhance the discrimination of attribute feature learning in space projection. Since we propose the identity-constraint attribute loss to make features contain both semantic attribute and identity information. Experimental results show the proposed method is robust to face attribute evaluation performance, and even applicable in other related tasks.

\subsection{TransFA Architecture}

\begin{figure*}[h]
    \centering
    \includegraphics[width=1\textwidth]{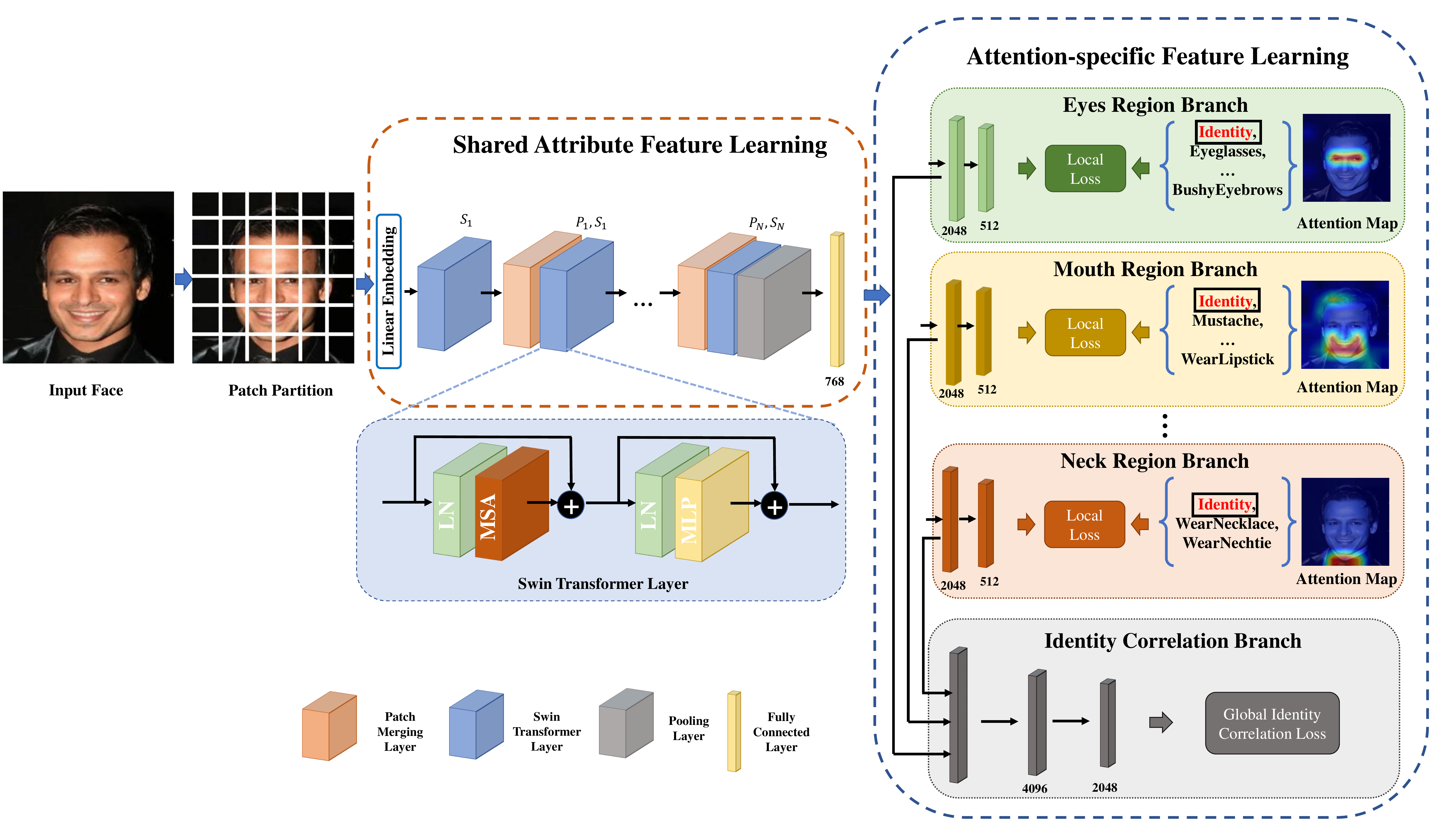}
    \caption{Overview of the proposed transformer-based representation for face attribute evaluation.}
    \label{fig:framework}
\end{figure*}

In this section, we give detailed descriptions about the proposed TransFA method for face attribute estimation task.
Figure \ref{fig:framework} shows the framework of the proposed algorithm in the training stage. The proposed TransFA consists of two modules: shared attribute feature learning module and attention-specific feature learning module. Considering the transformer could effectively model cross-region interactions, the shared attribute feature learning module mainly utilizes the Swin Transformer Layer \cite{Swin} to explore the strong inter-correlations between different attributes. Given the input face image, the regular window partition strategy \cite{ViT} is utilized to process input image into non-overleaping local windows. The patch size of each local window is $4\times4$. Then, a linear embedding layer maps it into 96-dim features. In the following Swin Transformer Layer, the standard multi-head self-attention \cite{ViT} (MSA) is used to process input firstly, and the designed multi-layer perceptron (MLP) is utilized to boost the transformation ability. Here the MLP consists of two fully-connected layers with GELU activation function, and a LayerNorm layers is added before them. 
As shown in Figure \ref{fig:framework}, given the input local window feature $p$, the detailed process of the Swin Transformer Layer is formulated as:
\begin{equation}
p' = MSA(LN(p)) + p,
\end{equation}
\begin{equation}
\hat p = MLP(LN(p')) + p'.
\end{equation}

In addition, we add several semantic attention-specific regions branches to evaluate strong-correlation face attribute based on the mentioned interpretable attribute grouping strategy. Inspired by the visualization of feature attention map of different attribute, we naturally group attributes with similar attention regions into the same category. Here we give an example of three attributes in one group (narrow eyes, bushy eyebrows and arched eyebrows) as shown in Figure 1. 
The attention maps of these attributes are rather similar, and naturally these are divided into the same category. 
Following, we utilize the proposed semi-automatic grouping strategy to divide all face attributes into seven attention-specific attribute groups in the following experiments. Details are shown in Table 1.

For further integrating the inherent relationship between semantic attributes and face identity, we concatenate the different attribute attention-specific features to obtain the global identity correlation feature. Experimental analysis proves the effectiveness of the proposed method in Section 4.4.

\subsection{Loss Function}

\textbf{Local attention-specific attribute loss:}
Considering input face images $X=\{x_{i}\}_{i=1}^N$, $N$ is the number of face image. 
The annotated face attributes are denoted by $Y=\{\{y_{i}^{a}\}_{a=1}^{A}\}_{i=1}^{N}$, $A$ is the number of face attributes. 
Here we consider the attribute evaluation task as classification task, and the binary cross-entropy function is chosen as the loss function:
\begin{equation}
    Loss_{A}=\sum_{a=1}^{A}\sum_{i=1}^{N}y_{i}^{a}log (p_{a} (x_{i},\theta_{a}))+ (1-y_{i}^{a})log (1-p_{a} (x_{i},\theta_{a})),
\end{equation}
here $p_{a} (x_{i},\theta_{a})$ denotes the prediction probability of data $x_{i}$ for the $a$-th attribute with model parameters $\theta_{a}$. 
Further, by considering the proposed interpretable attributes grouping strategy, the objective formulation could be rewritten as :
\begin{equation}
    \begin{aligned}
        Loss_{A}=& \sum_{g=1}^{G} \sum_{a=1}^{A_{g}}\sum_{i=1}^{N}y_{i}^{a}log (p_{a} (x_{i},\theta_{a}^{g}))+\\
        &(1-y_{i}^{a})log (1-p_{a} (x_{i},\theta_{a}^{g})).
    \end{aligned}
\end{equation}

Here $G$ means the number of attribute groups, $A_{g}$ is the number of attributes in the corresponding attribute group, and $\theta_{a}^{g}$ denotes model parameters evaluating the $a$-th attribute in group $g$.

Inspired the inherent identity-attribute relationship as described in Section 3.1, 
we assume the strong semantic discriminative representation could extract both attribute-related and identity-related information.
In other words, high-level identity information could be integrated to enhance semantic attribute evaluation performance.
Thus, the local identity loss is designed as follows:

\begin{equation}
    \begin{aligned}
    &Loss_{g} = \frac{1}{N (N-1)}\sum_{i=1}^{N}\sum_{j=i+1}^{N}w_{i,j}||fea_{i}^{g}-fea_{j}^{g}||_{2}^{2},
\end{aligned}
\end{equation}    
where 
\begin{equation}
    \begin{aligned}
&w_{i,j}=\begin{cases}
                1, \text{if $sample_{i}$ and $sample_{j}$ have the same identity},\\
                0,otherwise.
            \end{cases}
\end{aligned}
\end{equation}

Consequently, the local attention-specific attribute loss can be formulated as: 
\begin{equation}
    Loss_{LA} = \lambda Loss_{A} + \beta \sum_{g=1}^{G}Loss_{g},
\end{equation}
where $\lambda$ and $\beta$ are hyper-parameters which could balance the importance of the attribute discriminability influence and the identity discriminability influence.

\textbf{Global identity correlation loss:}
Due to further integrating the consistency between semantic attributes and face identity in the multi-scale perspective, we firstly concatenate diverse attention-specific attribute features to construct global attribute feature. And then different global face identity losses are introduced to enhance feature identity discriminability from different perspectives.
The common cross-entropy loss is chosen as follows:
\begin{equation}
    Loss_{F} = -\sum_{n=1}^{N}\sum_{c=1}^{C}log\frac{e^{p_{c} (x_{n})}}{\sum_{i=1}^{C}e^{p_{i} (x_{n})}}y_{n}^{c},
\end{equation}
where $C$ is the number of the identity in the training set,
$p_{i} (x_{n})$ is the prediction probability of the input $x_{n}$ of the $k$-th identity, $y_{n}^{c}$ means the identity label whose value is 1 when input $x_{n}$’s identity is the $k$-th subject.

Additionally, inspired by the wide application of embedding model in face recognition \cite{sun2014deep,ijcai18,TNNLS20}, we introduce another global identity loss as follows: 
\begin{equation}
\begin{aligned}
    &Loss_{C} = \frac{1}{N (N-1)}\sum_{i=1}^{N}\sum_{j=i+1}^{N}w_{i,j}||fea_{i}^{I}-fea_{j}^{I}||_{2}^{2},
\end{aligned}
\end{equation}
where
\begin{equation}
\begin{aligned}
    &w_{i,j}=\begin{cases}
                1, \text{if $sample_{i}$ and $sample_{j}$ have same identity},\\
                0,otherwise.
            \end{cases}
\end{aligned}
\end{equation}

Therefore, the complete global identity correlation loss can be formulated as:
\begin{equation}
    Loss_{GI} = \alpha Loss_{F} + (1-\alpha)Loss_{C}.
\end{equation}

\textbf{Hierarchical identity-constraint attribute loss:}
For considering the relationship between semantic attributes and face identity in the multi-scale perspective, these mentioned two different losses are combined linearly finally.
Thus, the proposed hierarchical identity-constraint attribute loss can be formulated as:
\begin{equation}
    Loss_{total} = Loss_{GI} + Loss_{LA}.
\end{equation}

\subsection{Implementation Details} \label{Implementation Details}

The goal of our method is to evaluate multiple face attributes simultaneously by exploiting the relationship among them.
Most former attribute evaluation methods \cite{han2017heterogeneous, mao2020deep} prove the multi-task learning strategy would be suitable in this task.
However, how to group these diverse face attributes reasonably is still a challenge. 
Inspired by the visualization of attention maps (shown in Figure \ref{fig:CNN&Trans}) by transformer-based methods, we naturally select face attributes with similar attention regions into the same group.
Thus, attributes in the global group such as attractive, young, and chubby describe the whole face characteristic, while attributes in the local group such as big nose, bald, and bangs, mainly portray the character of the local facial components.
Different from prior methods grouping face attribute always depending on prior knowledge \cite{han2017heterogeneous, mao2020deep} (like data type, semantic meaning and subjectivity),
the proposed  \emph{attention-specific attribute grouping strategy} could effectively divide face attributes into suitable categories, which could further enhance the inherent spatial inter-correlations. 
The grouping details are shown in table \ref{tab:attributes_grouping}.


In addition, the shared attribute feature learning module is a modified Swin Transformer \cite{Swin} (12 Swin Transformer Layer, 10 patch merging layers) with a BatchNorm layer and a pooling layer behind. 
The patch merging layer could help reduce the number of feature dimensions. 
For further enhance connections across local windows, we follow the same shifted window partition with \cite{Swin}. Thus, the shifted window partitioning is to shift features by $2\times2$ pixels.
Here we alternately utilize the regular and shifted windows partitioning in the Swin Transformer Layer.
Each attention-specific region branch network consists of three fully-connected layers with ReLU non-linearity and dropout operators inserted. 

We perform stochastic gradient descent (SGD) to optimize parameters in the proposed shared attribute feature learning module and attention-specific feature learning module in an end-to-end way. 
The start learning rate starts from 0.01 and it drops by 10 every 5 epochs. We use a momentum of 0.9 here.
The proposed TransFA related experiments are conducted on NVIDIA GeForce RTX 3090 GPU. 

\begin{table}[t]
\caption{Facial attributes and their corresponding groupings.}
\label{tab:attributes_grouping}
\begin{tabular}{ll}
\hline
\textbf{Group}                        & \textbf{Attributes }                                                                                                                  \\\hline
\multirow{2}{*}{\centering \textbf{Global}}    & \multirow{2}{0.33\textwidth}{Attractive, Blurry, Chubby, Heavy Makeup, Male, Oval Face, Pale Skin, Smiling, Young}                           \\
                             &                                                                                                                              \\\hline
\multirow{3}{*}{\centering \textbf{Around head}} & \multirow{3}{ 0.33\textwidth}{Bald, Bangs, Black Hair, Blond Hair, Brown Hair, Gray Hair, Receding Hairline, Straight Hair, Wavy Hair, Wear Hat}   \\
                             &                                                                                                                              \\
                             &                                                                                                                              \\\hline
\multirow{2}{*}{\centering \textbf{Eyes}}        & \multirow{2}{ 0.33\textwidth}{Arched Eyebrows, Bags Under Eyes, Bushy Eyebrows, Eyeglasses, Narrow Eyes}                                        \\
                             &                                                                                                                              \\\hline
\textbf{Nose}                         & Big Nose, Pointy Nose                                                                                                          \\\hline
\multirow{3}{*}{\centering \textbf{Mouth}}       & \multirow{3}{ 0.33\textwidth}{5 O’Clock Shadow, Big Lips, Double Chin, Goatee, Mouth Slightly Open, Mustache, No Beard, Sideburns, Wear Lipstick} \\
                             &                                                                                                                              \\
                             &                                                                                                                              \\\hline
\multirow{2}{*}{\centering \textbf{Cheeks}}                       & \multirow{2}{ 0.33\textwidth}{High Cheekbones, Rosy Cheeks, Wear Earrings}\\ &                                                                                     \\\hline
\textbf{Neck}                         & Wear Necklace, Wear Necktie            \\ \hline                                                                                       
\end{tabular}
\end{table}
\section{EXPERIMENTS} \label{EXPERIMENTS}
In this section, we present experiments on two representative face attribute databases to verify the effectiveness of our method. We first briefly introduce these two databases and our experiment settings. Then, we compare our method with state-of-art approaches. Finally, we perform the ablation study to investigate the effectiveness of each component of our method and analyze how our algorithm perform beyond pure attribute evaluation task.
\subsection{Datasets}
\begin{figure}
    \centering
    \includegraphics[width=0.49\textwidth]{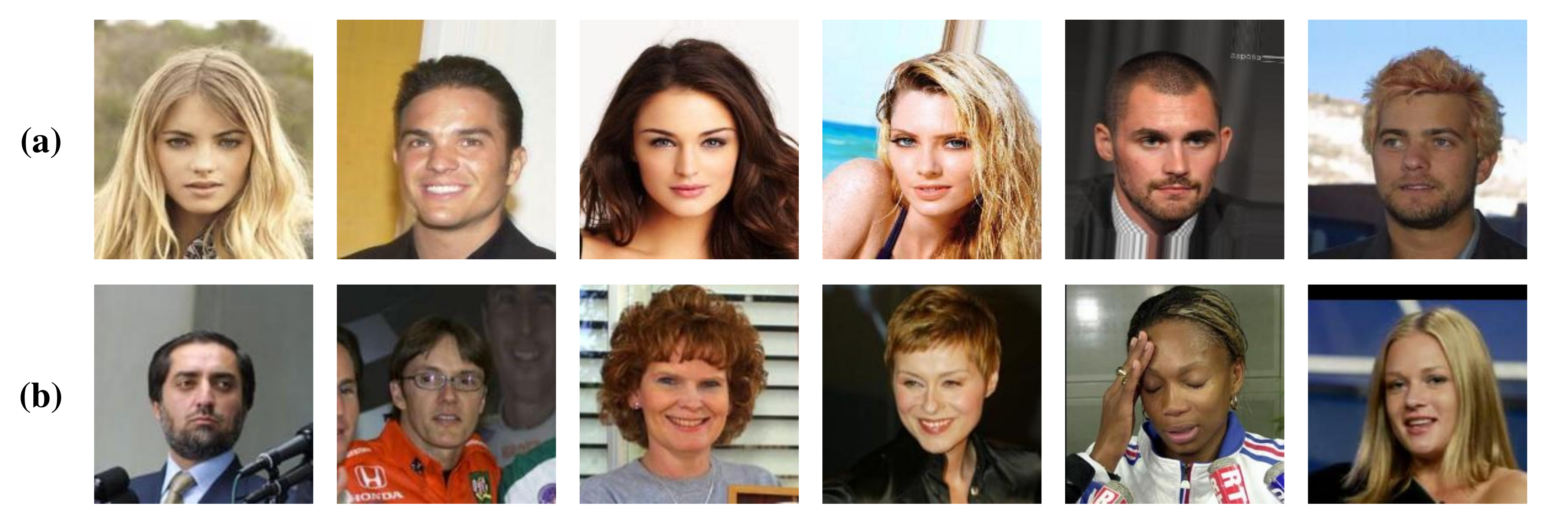}
    \caption{The illustration of face attribute databases. (a): CelebA database (b): LFWA database}
    \label{fig:dataset}
\end{figure}

In our experiments, we use two challenging public face attribute databases for evaluation: Large-scale CelebFaces Attributes (CelebA) Database \cite{liu2015deep} and Labeled Faces in the Wild Attributes (LFWA) Database \cite{liu2015deep}. Example face images are shown in Figure \ref{fig:dataset}.
\begin{itemize}
    \item \textbf{CelebA} \cite{liu2015deep} is a public large-scale face attribute database consisting of 200K images from more than 10K person, and each image contains 40 face attribute annotations, making it challenging to evaluate all face attributes simultaneously. The 200K images in CelebA are divided into training set, validation set, and test set with the size of 162,770, 19,867, and 19,962, For fair comparison, we follow the same protocol \cite{liu2015deep} in experiments.

    \item \textbf{LFWA} \cite{liu2015deep} is also a large-scale face attribute database that contains the same images available in LFW \cite{huang2008labeled}, and each image contains 40 face attribute annotations. This database totally includes 13,143 images, and we follow the same protocol \cite{liu2015deep}. The 6263 images are selected for training and the rest 6880 images are chosen for testing.
\end{itemize}

\subsection{Experimental Settings}
\begin{figure}
    \centering
    \begin{subfigure}[b]{0.45\linewidth}
        \centering
        \includegraphics[scale=0.27]{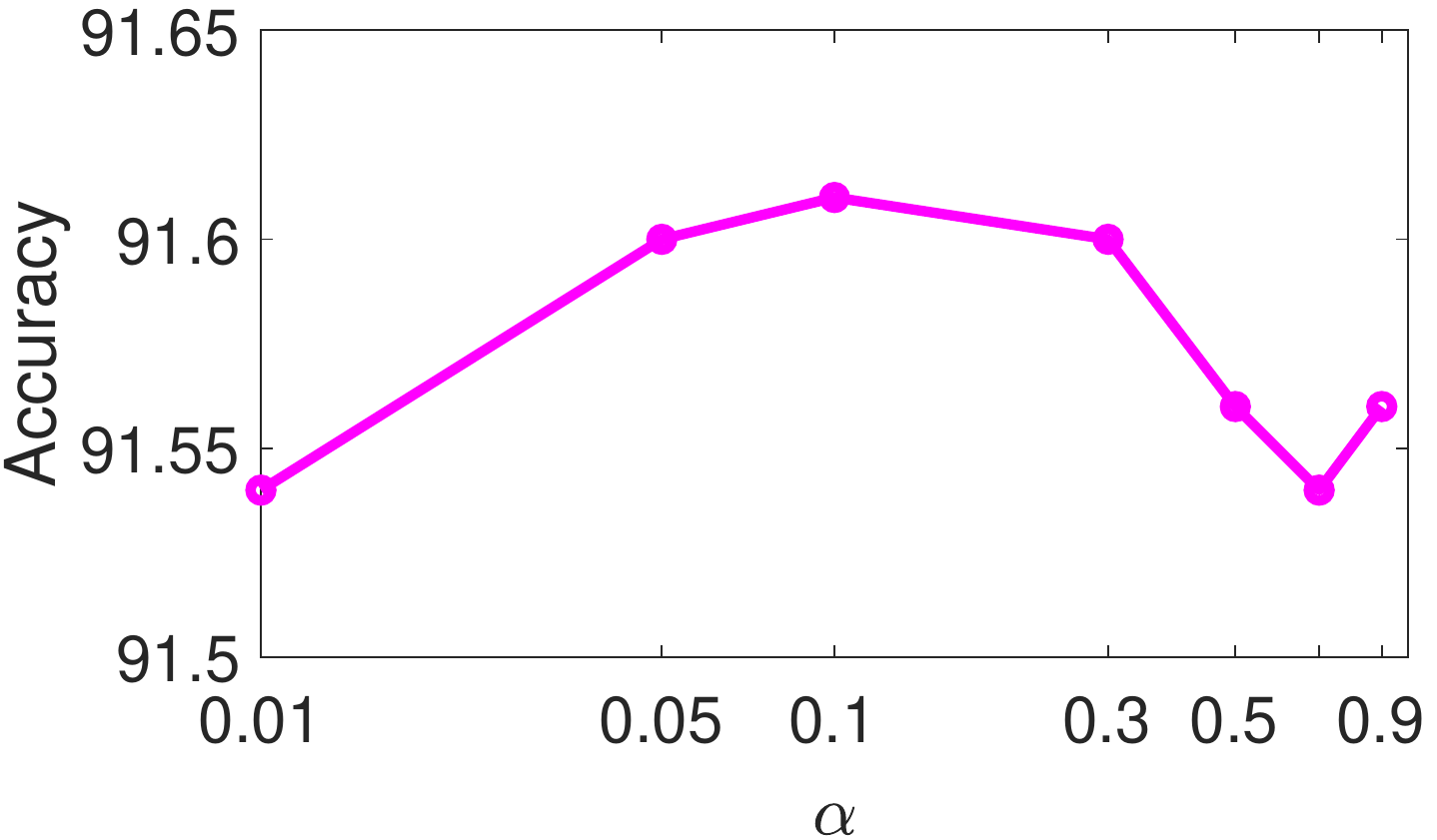}
    \end{subfigure}
    \hspace{0mm}
    \begin{subfigure}[b]{0.45\linewidth}
        \centering
        \includegraphics[scale=0.27]{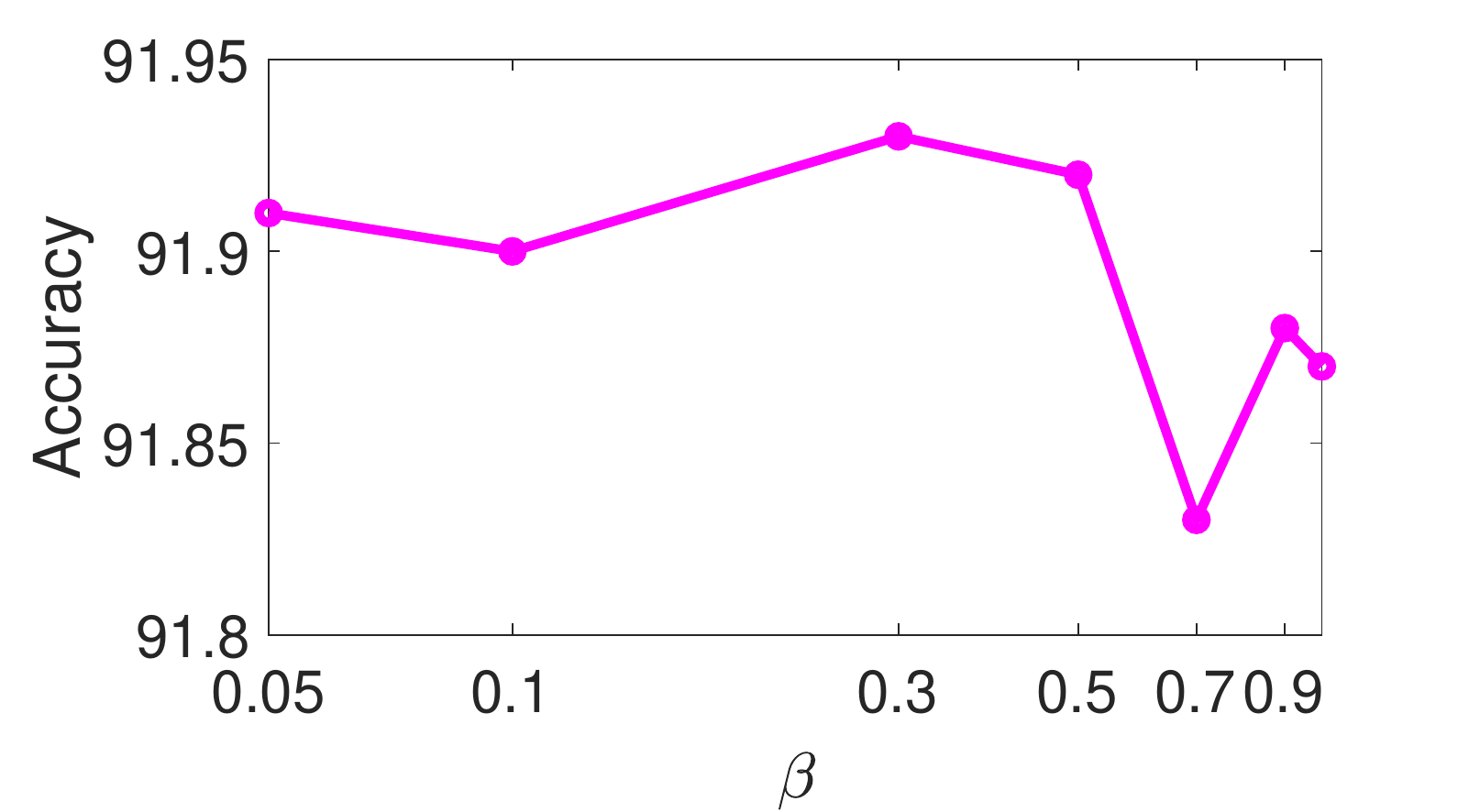}
    \end{subfigure}
    \caption{The left subfigure shows the evaluation accuracies (in \%) of different numbers of the parameter $\alpha$ on CelebA database; the right subfigure presents the evaluation accuracies (in \%) of different numbers of the parameter $\beta$ on CelebA database}
    \label{fig: hyperparameters}
\end{figure}

\textbf{Preprocessing} 
 We first calculate the number of identities in the training set and construct the identity correlation branch based on the number of identities. Then we divide facial attributes into seven groups based on the proposed semi-automatic attribute grouping strategy. Next, we rearrange the order of the attribute annotations of the public attribute database. 
 Finally, we normalize all the face image mean [0.5, 0.5, 0.5] and standard deviation [0.5, 0.5, 0.5], and finally resize them to a fixed size of 224 $\times$ 224.

\textbf{Influence of parameter $\alpha$} To evaluate the effect of parameter $\alpha$ on the evaluation performance, parameters of different sizes are selected to conduct experiments on CelebA datasets. In the left subfigure of Figure \ref{fig: hyperparameters}, we illustrate the effect of parameter $\alpha$ of different values from a set of \{ 0.05, 0.10, 0.30, 0.50, 0.70, 0.90 \} when the parameter $\beta$ is set at fixed value 0. We find that the evaluation accuracy varies with different $\alpha$, and the performance is better when the value of parameter $\alpha$ is approximately 0.1.

\textbf{Influence of parameter $\beta$} We also investigated the effect of another parameter $\beta$ on CelebA. In the right subfigure of Figure \ref{fig: hyperparameters}, we demonstrate the effect of parameter $\beta$ from a set of \{0.05, 0.10, 0.30, 0.50, 0.70, 0.90, 1.00\} when parameter $\alpha$ is set at a fixed value of 0.1. After analyzing the experimental results, we find that when $\beta$ reaches nearly 0.3 leading better performance.
The parameter $\alpha$ balances the importance of global identity correlation branch and $\beta$  represents the importance of the local attribute attention-specific branch. After conducting experiments on different parameters, we decided to set $\alpha$ as 0.1 and set $\beta$  as 0.3 in the formulation of proposed hierarchical identity-constraint attribute loss.
The parameter $\lambda$ is set as a fixed value of 5 from experience.

\subsection{Comparison Results}
\label{Comparison with State-of-Art Methods}

\begin{table*}[]
\caption{The attribute evaluation accuracies (in \%) of 40 facial attributes on CelebA and LFWA by several state-of-the-art methods \cite{kumar2008facetracer}, \cite{zhang2014panda}, \cite{liu2015deep}, \cite{hand2017attributes}, \cite{mahbub2018segment}, \cite{mao2020deep}, \cite{han2017heterogeneous} and TransFA with the highest result is bolded and the second-highest result is underlined on each attribute. Besides, the average accuracies of \cite{kumar2008facetracer}, \cite{zhang2014panda}, \cite{liu2015deep}, \cite{hand2017attributes}, \cite{mahbub2018segment}, \cite{mao2020deep}, \cite{han2017heterogeneous} and TransFA on two databases are shown below: 82.55\%, 85.43\%, 87.33\%, 91.29\%, 90.61\%, 91.70\%, 92.60\%, and 91.93\% on CelebA; 73.93\%, 81.03\%, 83.85\%, 86.31\%, 85.82\%, 86.56\%, 86.15\%, and 86.57\% on LFWA.
}
\label{tab:result}
\resizebox{\textwidth}{!}{
\begin{tabular}{llcccccccccccccccccccc}
\hline
\multicolumn{2}{c}{\multirow{2}{*}{\textbf{Approach}}} & \multicolumn{20}{c}{\textbf{Attribute index}}                                                                                                                                                                                                                                                                                                              \\ \cline{3-22} 
\multicolumn{2}{l}{}                          & 1              & 2              & 3              & 4              & 5              & 6              & 7              & 8              & 9              & 10             & 11             & 12             & 13             & 14             & 15             & 16             & 17             & 18             & 19             & 20             \\\hline
\multirow{8}{*}{\centering\textbf{CelebA}}      & FaceTracer \cite{kumar2008facetracer} & 85.00          & 76.00          & 78.00          & 76.00          & 89.00          & 88.00          & 64.00          & 74.00          & 85.00          & 93.00          & 81.00          & 77.00          & 86.00          & 86.00          & 88.00          & 98.00          & 93.00          & 90.00          & 85.00          & 84.00          \\
 & PANDA \cite{zhang2014panda}      & 88.00          & 78.00          & 81.00          & 79.00          & 96.00          & 92.00          & 67.00          & 75.00          & 85.00          & 93.00          & 86.00          & 77.00          & 86.00          & 86.00          & 88.00          & 98.00          & 93.00          & 94.00          & 90.00          & 86.00          \\
 & LNets+ANet \cite{liu2015deep} & 91.00          & 79.00          & 81.00          & 79.00          & 98.00          & 95.00          & 68.00          & 78.00          & 88.00          & 95.00          & 84.00          & 80.00          & 90.00          & 91.00          & 92.00          & 99.00          & 95.00          & 97.00          & 90.00          & {\ul 88.00}    \\
 & MCNN-AUX \cite{hand2017attributes}        & 94.51          & 83.42          & 83.06          & 84.92          & 98.90          & 96.05          & 71.47          & 84.53          & 89.78          & 96.01          & 96.17          & 89.15          & 92.84          & 95.67          & 96.32          & 99.63          & 97.24          & 98.20          & 91.55          & 87.58          \\
 & NSA \cite{mahbub2018segment}       & 93.13          & 82.56          & 82.76          & 84.86          & 98.03          & 95.71          & 69.28          & 83.81          & 89.03          & 95.76          & 95.96          & 88.25          & 92.66          & 94.94          & 95.80          & 99.51          & 96.68          & 97.45          & 91.59          & 87.61          \\
 & DMM-CNN \cite{mao2020deep}   & 94.84          & 84.57          & {\ul 83.37}    & 85.81          & {\ul 99.03}    & \textbf{96.22} & {\ul 72.93}    & 84.78          & {\ul 90.50}    & {\ul 96.13}    & {\ul 96.40}    & {\ul 89.46}    & {\ul 93.01}    & {\ul 95.86}    & 96.39          & \textbf{99.69} & 97.63          & {\ul 98.27}    & 91.85          & 87.73          \\
 & DMTL \cite{han2017heterogeneous}       & {\ul 95.00}    & \textbf{86.00} & \textbf{85.00} & \textbf{99.00} & 99.00          & 96.00          & \textbf{88.00} & \textbf{92.00} & 85.00          & 91.00          & 96.00          & \textbf{96.00} & 85.00          & \textbf{97.00} & \textbf{99.00} & 99.00          & \textbf{98.00} & 96.00          & {\ul 92.00}    & {\ul 88.00}    \\
 & TransFA    & \textbf{95.24} & {\ul 84.78}    & 83.25          & {\ul 86.07}    & \textbf{99.05} & {\ul 96.18}    & 72.58          & {\ul 85.55}    & \textbf{90.80} & \textbf{96.16} & \textbf{96.51} & 89.14          & \textbf{93.03} & 95.79          & {\ul 96.63}    & {\ul 99.67}    & {\ul 97.70}    & \textbf{98.35} & \textbf{92.19} & \textbf{88.29} \\ \hline
\multirow{8}{*}{\centering\textbf{LFWA}}        & FaceTracer \cite{kumar2008facetracer} & 70.00          & 67.00          & 71.00          & 65.00          & 77.00          & 72.00          & 68.00          & 73.00          & 76.00          & 88.00          & 73.00          & 62.00          & 67.00          & 67.00          & 70.00          & 90.00          & 69.00          & 78.00          & 88.00                               & 77.00          \\
                             & PANDA \cite{zhang2014panda}   & \textbf{84.00} & 79.00          & 81.00          & 80.00          & 84.00          & 84.00          & 73.00          & 79.00          & 87.00          & 94.00          & 74.00          & 74.00          & 79.00          & 69.00          & 75.00          & 89.00          & 75.00          & 81.00          & 93.00                               & 86.00          \\
                             & LNets+ANet \cite{liu2015deep}    & \textbf{84.00} & 82.00          & {\ul 83.00}    & 83.00          & 88.00          & 88.00          & 75.00          & 81.00          & 90.00          & 97.00          & 74.00          & 77.00          & 82.00          & 73.00          & 78.00          & \textbf{95.00} & 78.00          & 84.00          & 95.00                               & 88.00          \\
                             & MCNN-AUX \cite{hand2017attributes}    & 77.06          & 81.78          & 80.31          & {\ul 83.48}    & 91.94          & 90.08          & 79.24          & \textbf{84.98} & \textbf{92.63} & {\ul 97.41}    & 85.23          & 80.85          & 84.97          & 76.86          & 81.52          & 91.30          & 82.97          & 88.93          & \textbf{95.85}                      & 88.38          \\
                             & NSA \cite{mahbub2018segment}  & 77.59          & 81.72          & 80.16          & 82.62          & 91.88          & {\ul 90.71}    & 78.97          & 83.13          & {\ul 92.49}    & \textbf{97.47} & 86.42          & 80.93          & 84.26          & 76.06          & 80.49          & 91.50          & 83.01          & 88.46          & 95.39                               & 88.34          \\
                             & DMM-CNN \cite{mao2020deep}  & 79.18          & 82.70          & 81.10          & 82.70          & 91.96          & \textbf{91.30} & 79.82          & {\ul 83.67}    & 91.55          & 97.17          & \textbf{87.58} & 81.56          & \textbf{85.33} & {\ul 77.66}    & 80.98          & {\ul 92.83}    & 82.82          & \textbf{89.38} & {\ul 95.68}                         & 88.13          \\
                             & DMTL \cite{han2017heterogeneous}  & {\ul 80.00}    & \textbf{86.00} & \textbf{84.00} & \textbf{92.00} & \textbf{93.00} & 77.00          & \textbf{81.00} & 80.00          & 83.00          & 92.00          & 75.00          & \textbf{97.00} & 82.00          & \textbf{78.00} & \textbf{92.00} & 86.00          & \textbf{88.00} & {\ul 89.00}    & 95.00                               & {\ul 89.00}    \\
                             & TransFA       & 79.35          & {\ul 83.02}    & 80.26          & 83.26          & {\ul 92.11}    & 90.57          & {\ul 80.26}    & 83.26          & 92.33          & 96.89          & {\ul 86.88}    & {\ul 82.27}    & {\ul 85.32}    & 77.51          & {\ul 82.12}    & 92.38          & {\ul 83.82}    & 88.94          & 95.45                               & \textbf{89.04} \\ \hline
\multicolumn{2}{c}{\multirow{2}{*}{\textbf{Approach}}} & \multicolumn{20}{c}{\textbf{Attribute index}}                                                                                                                                                                                                                                                                                                              \\ \cline{3-22} 
\multicolumn{2}{l}{}                          & 21             & 22             & 23             & 24             & 25             & 26             & 27             & 28             & 29             & 30             & 31             & 32             & 33             & 34             & 35             & 36             & 37             & 38             & 39             & 40             \\ \hline
\multirow{8}{*}{\centering\textbf{CelebA}}      & FaceTracer \cite{kumar2008facetracer} & 91.00          & 87.00          & 91.00          & 82.00          & 90.00          & 64.00          & 83.00          & 68.00          & 76.00          & 84.00          & 94.00          & 89.00          & 69.00          & 73.00          & 73.00          & 89.00          & 89.00          & 68.00          & 86.00          & 80.00          \\
 & PANDA \cite{zhang2014panda}       & 97.00          & 93.00          & 93.00          & 84.00          & 93.00          & 65.00          & 91.00          & 71.00          & 85.00          & 87.00          & 93.00          & 92.00          & 69.00          & 77.00          & 78.00          & 96.00          & 93.00          & 67.00          & 91.00          & 84.00          \\
 & LNets+ANet \cite{liu2015deep} & 98.00          & 92.00          & 95.00          & 81.00          & 95.00          & 66.00          & 91.00          & 72.00          & 89.00          & 90.00          & 96.00          & 92.00          & 73.00          & 80.00          & 82.00          & 99.00          & 93.00          & 71.00          & 93.00          & 87.00          \\
 & MCNN-AUX \cite{hand2017attributes}    & 98.17          & 93.74          & 96.88          & 87.23          & 96.05          & 75.84          & {\ul 97.05}    & 77.47          & 93.81          & 95.16          & 97.85          & 92.73          & 83.58          & 83.91          & 90.43          & 99.05          & 94.11          & 86.63          & 96.51          & 88.48          \\
 & NSA \cite{mahbub2018segment}        & 97.95          & 93.78          & 95.86          & 86.88          & 96.17          & 74.93          & 97.00          & 76.47          & 92.25          & 94.79          & 97.17          & 92.70          & 80.41          & 81.70          & 89.44          & 98.74          & 93.21          & 85.61          & 96.05          & 88.01          \\
 & DMM-CNN \cite{mao2020deep}    & {\ul 98.29}    & {\ul 94.16}    & {\ul 97.03}    & 87.73          & 96.41          & 75.89          & 97.00          & 77.19          & {\ul 94.12}    & 95.32          & 97.91          & 93.22          & 84.72          & 86.01          & 90.78          & {\ul 99.12}    & {\ul 94.49}    & 88.03          & {\ul 97.15}    & 88.98          \\
 & DMTL \cite{han2017heterogeneous}       & 98.00          & 94.00          & 97.00          & \textbf{90.00} & \textbf{97.00} & \textbf{78.00} & 97.00          & {\ul 78.00}    & 94.00          & \textbf{96.00} & {\ul 98.00}    & \textbf{94.00} & {\ul 85.00}    & \textbf{87.00} & \textbf{91.00} & 99.00          & 93.00          & \textbf{89.00} & 97.00          & \textbf{90.00} \\
 & TransFA    & \textbf{98.85} & \textbf{94.24} & \textbf{97.18} & {\ul 87.95}    & {\ul 96.56}    & {\ul 77.17}    & \textbf{97.27} & \textbf{78.18} & \textbf{94.13} & {\ul 95.40}    & \textbf{98.04} & {\ul 93.66}    & \textbf{85.22} & {\ul 86.31}    & {\ul 90.80}    & \textbf{99.22} & \textbf{94.65} & {\ul 88.34}    & \textbf{97.17} & {\ul 89.73}   
    \\ \hline
\multirow{8}{*}{\centering\textbf{LFWA}}        & FaceTracer \cite{kumar2008facetracer}   & 84.00          & 77.00          & 83.00          & 73.00          & 69.00          & 66.00          & 70.00          & 74.00          & 63.00          & 70.00          & 71.00          & 78.00          & 67.00          & 62.00          & 88.00          & 75.00          & 87.00          & 81.00          & 71.00                               & 80.00          \\
                             & PANDA \cite{zhang2014panda}  & 92.00          & 78.00          & 87.00          & 73.00          & 75.00          & 72.00          & 84.00          & 76.00          & 84.00          & 73.00          & 76.00          & 89.00          & 73.00          & 75.00          & 92.00          & 82.00          & 93.00          & 86.00          & 79.00                               & 82.00          \\
                             & LNets+ANet \cite{liu2015deep} & 94.00          & 82.00          & 92.00          & 81.00          & 79.00          & 74.00          & 84.00          & 80.00          & 85.00          & 78.00          & 77.00          & 91.00          & 76.00          & 76.00          & 94.00          & 88.00          & 95.00          & 88.00          & 79.00                               & 86.00          \\
                             & MCNN-AUX \cite{hand2017attributes}   & {\ul 94.02}    & 83.51          & 93.43          & 82.86          & 82.15          & {\ul 77.39}    & \textbf{93.32} & 84.14          & 86.25          & \textbf{87.92} & \textbf{83.13} & 91.83          & 78.53          & \textbf{81.61} & \textbf{94.95} & 90.07          & {\ul 95.04}    & 89.94          & 80.66                               & 85.84          \\
                             & NSA \cite{mahbub2018segment}    & 92.60          & 82.50          & 92.97          & 82.75          & 80.77          & 76.80          & 90.97          & 84.20          & 84.90          & {\ul 87.08}    & 81.76          & 90.80          & 78.91          & 78.28          & {\ul 94.75}    & 90.23          & 94.07          & 89.59          & \textbf{81.40}                      & 85.68          \\
                             & DMM-CNN \cite{mao2020deep}  & \textbf{94.14} & {\ul 84.45}    & {\ul 94.46}    & {\ul 83.67}    & {\ul 82.48}    & 76.94          & {\ul 91.86}    & {\ul 84.51}    & \textbf{86.30} & 86.44          & {\ul 82.99}    & \textbf{92.24} & {\ul 79.20}    & 79.87          & 94.14          & {\ul 90.84}    & \textbf{95.11} & 89.47          & {\ul 81.28}                         & \textbf{88.94} \\
                             & DMTL \cite{han2017heterogeneous}   & 93.00          & \textbf{86.00} & \textbf{95.00} & 82.00          & 81.00          & 75.00          & 91.00          & 84.00          & 85.00          & 86.00          & 80.00          & {\ul 92.00}    & 79.00          & 80.00          & 94.00          & \textbf{92.00} & 93.00          & \textbf{91.00} & 81.00                               & {\ul 87.00}    \\
                             & TransFA   & 93.79          & 83.28          & 94.13          & \textbf{83.91} & \textbf{82.66} & \textbf{77.60} & 91.50          & \textbf{84.56} & {\ul 86.29}    & \textbf{87.92} & 82.21          & 91.25          & \textbf{80.70} & {\ul 80.28}    & 94.53          & 90.51          & 94.69          & {\ul 90.07}    & 81.25                               & 86.69         \\ \hline     
\end{tabular}}
\end{table*}

To verify the effectiveness of the proposed TransFA, we compare TransFA with other state-of-the-art methods \cite{kumar2008facetracer, zhang2014panda, liu2015deep, hand2017attributes, mahbub2018segment, mao2020deep, han2017heterogeneous} on CelebA and LFWA databases with the same protocol provided in \cite{liu2015deep}.
The comparison results are reported in Table \ref{tab:result}. 
It can be seen that our network obtained competitive performance with test accuracy of 91.93\% and 86.57\% on CelebA and LFWA dataset respectively.
The DMTL \cite{han2017heterogeneous} introduce extra face normaliztiton step to process input faces to eliminate the adverse impact of complex background, which it also increase the computational cost.
The proposed TransFA could achieve similar average accuracy on CelebA dataset, while image face are directly input the backbone model.
Besides, the proposed TransFA also could achieve the best accuracy for 20 attributes among 40 face attributes compared with the SOTA methods.
This is because the transformer-based representation could focus on more attribute discriminative regions.
Benefited by the increased training data scale, the proposed TransFA further achieves the superior performance in average recognition accuracy on LFWA dataset.

\begin{figure}[h]
\centering
\includegraphics[width=0.42\textwidth]{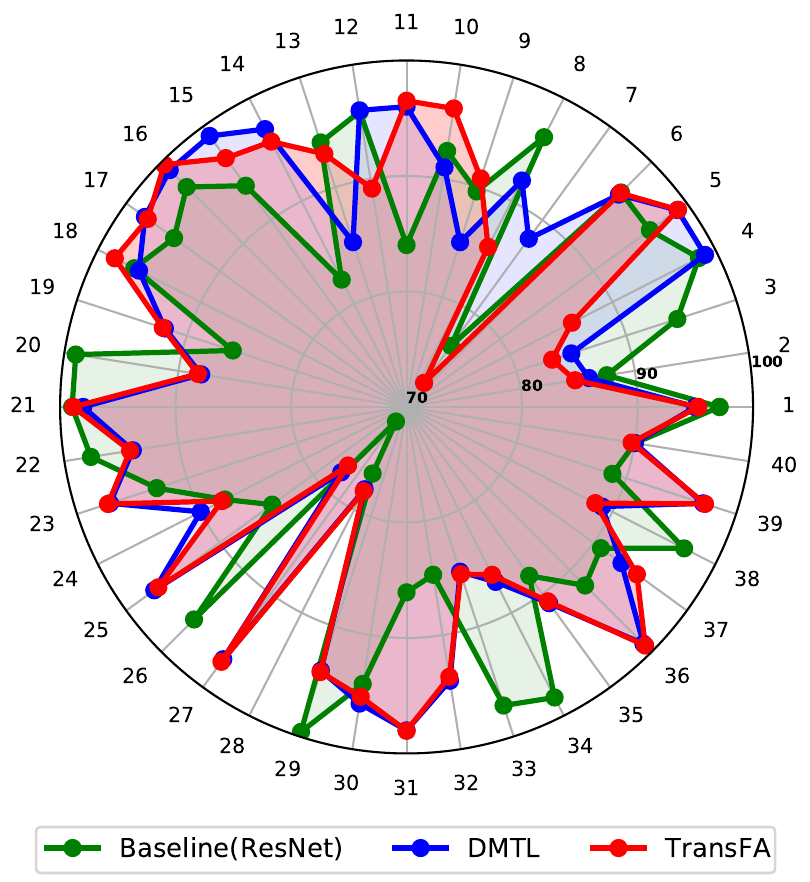}
\caption{Evaluation accuracies of different face attributes on CelebA dataset of Baseline(ResNet), DMTL \cite{han2017heterogeneous} and TransFA with 91.50\%, 92.60\% and 91.93\% as average accuracies respectively.}
\label{fig:radar}
\end{figure}

To demonstrate the effectiveness of proposed TransFA more intuitively, we compare three representative approaches in Figure \ref{fig:radar}, including Baseline (ResNet), DMTL \cite{han2017heterogeneous} and TransFA. 
It is noted that we use ResNet-50 Network as backbone for shared feature learning and incorporate the proposed attention-specific attribute grouping strategy for attribute evaluation on baseline experiment.
The pure multi-label cross-entropy loss is utilized in this multi-label classification task.
Compared with the baseline, TransFA outperforms it on the most attribute, and achieves 0.43\% higher than it on average accuracy. 
It could be ascribed to the effectiveness of transformer-based representation compared with CNN-based representation.
Additionally, the proposed TransFA achieves the best attribute accuracy on the most face attributes.
For example, the proposed TransFA could evaluate attributes bald (attribute index 9) and bangs (attribute index 10) at accuracy of 90.80\% and 96.16\% separately, which increases 5.80\% and 5.16\% compared with \cite{han2017heterogeneous}.
It strongly proves the proposed transformer-based representation could integrate the inter-correlation of face attributes more effectively.
However, our method still has a certain gap with it \cite{han2017heterogeneous} on evaluating attribute big lips (attribute index 27), which may results from too many attributes divided into the mouth group leading to the adverse effect.

\subsection{Algorithm Analysis}
\textbf{Ablation study}
Here we analyze each key component in the proposed TransFA in Table  \ref{tab:ablation}.
It is noted that the simple baseline experiment is only conducted with the pure multi-label cross entropy as objective function, and the ResNet-50 as the backbone network for convenience of comparison.
When utilizing the proposed attention-specific attribute grouping strategy, the average accuracy increase 0.44\% which proves the semi-automatic grouping could help enhance attributes correlations.
In the following experiment, the proposed transformer backbone replace CNN backbone in the shared attribute feature learning module, which could raise the evaluation accuracy to 91.57 \%.
Finally, the proposed full TransFA could further increase 0.36\% accuracy, which proves the proposed method could effectively integrate the relationship between semantic attributes and high-level identity information. 

\begin{table}[]
\caption{Ablation study of different components in the proposed TransFA on CelebA database (in \%). }
\begin{tabular}{cccc}
\hline
\multicolumn{1}{c}{\multirow{3}{*}{\textbf{\begin{tabular}[c]{@{}c@{}}Grouping\\ Strategy\end{tabular}}}} & \multicolumn{1}{c}{\multirow{3}{*}{\textbf{\begin{tabular}[c]{@{}c@{}}Transformer\\ Backbone\end{tabular}}}} & \multicolumn{1}{c}{\multirow{3}{*}{\textbf{\begin{tabular}[c]{@{}c@{}}Identity-\\ constraint\\ Attribute Loss\end{tabular}}}} & \multicolumn{1}{c}{\multirow{3}{*}{\textbf{\begin{tabular}[c]{@{}c@{}}Average\\ Accuracy\end{tabular}}}} \\
\multicolumn{1}{c}{}                                                                                      & \multicolumn{1}{c}{}                                                                                         & \multicolumn{1}{c}{}                                                                                                          & \multicolumn{1}{c}{}                                                                                  \\
\multicolumn{1}{c}{}                                                                                      & \multicolumn{1}{c}{}                                                                                         & \multicolumn{1}{c}{}                                                                                                          & \multicolumn{1}{c}{}                                                                                  \\\hline
    -        &       -        &   -    &    91.06                                   \\
    \checkmark          &       -        &   -    & 91.50                                      \\
    \checkmark          &       \checkmark          &   -    & 91.57                                        \\
    \checkmark          &       \checkmark          &   \checkmark      &   91.93            \\ \hline                                                                                        
\end{tabular}
\label{tab:ablation}
\end{table}

\textbf{Beyond attribute evaluation task}
To exploit the effectiveness of modeling the relationship of identity and attribute in our proposed method, we conduct experiments to explore the identity-related discriminative information in the proposed transformer-based representation in face recognition task.
We randomly selected 100 people from the test set of CelebA dataset with each person has two images at least. Then, one image from each identity is randomly selected to compose the probe set, and the rest images compose the gallery set. The results are demonstrated in Table \ref{tab:FR}. We can find that the feature learned from global identity correlation branch performs better than the features learned from local attribute attention-specific branch. All experiments have a acceptable Rank-1 accuracy, which demonstrates the proposed TransFA representation could effectively model the relationship of face attribute and identity. It also inspires researchers to explore the general face representation model in multiple face analysis tasks.
\begin{table}[!ht]
\caption{The face recognition accuracies (in\%) at Rank-1 with representations from different branches.}
\label{tab:FR}
\centering
\begin{tabular}{ll}
\hline
\textbf{Attention-specific Attribute Branch} & \textbf{Accuracy}  \\ \hline
        \textbf{Global Region Branch}  & 87.00  \\
        \textbf{Head Region Branch}  & 87.00  \\ 
        \textbf{Eyes Region Branch}  & 89.00 \\ 
        \textbf{Nose Region Branch}  & 88.00  \\ 
        \textbf{Mouth Region Branch}  & 88.00  \\ 
        \textbf{Cheeks Region Branch}  & 87.00  \\ 
        \textbf{Neck Region Branch}  & 90.00  \\ 
        \textbf{Identity Correlation Branch}  & 93.00  \\ \hline
    \end{tabular}
\end{table}

\textbf{Failure case}
\begin{figure}[t]
    \centering
    \includegraphics[width=0.45\textwidth]{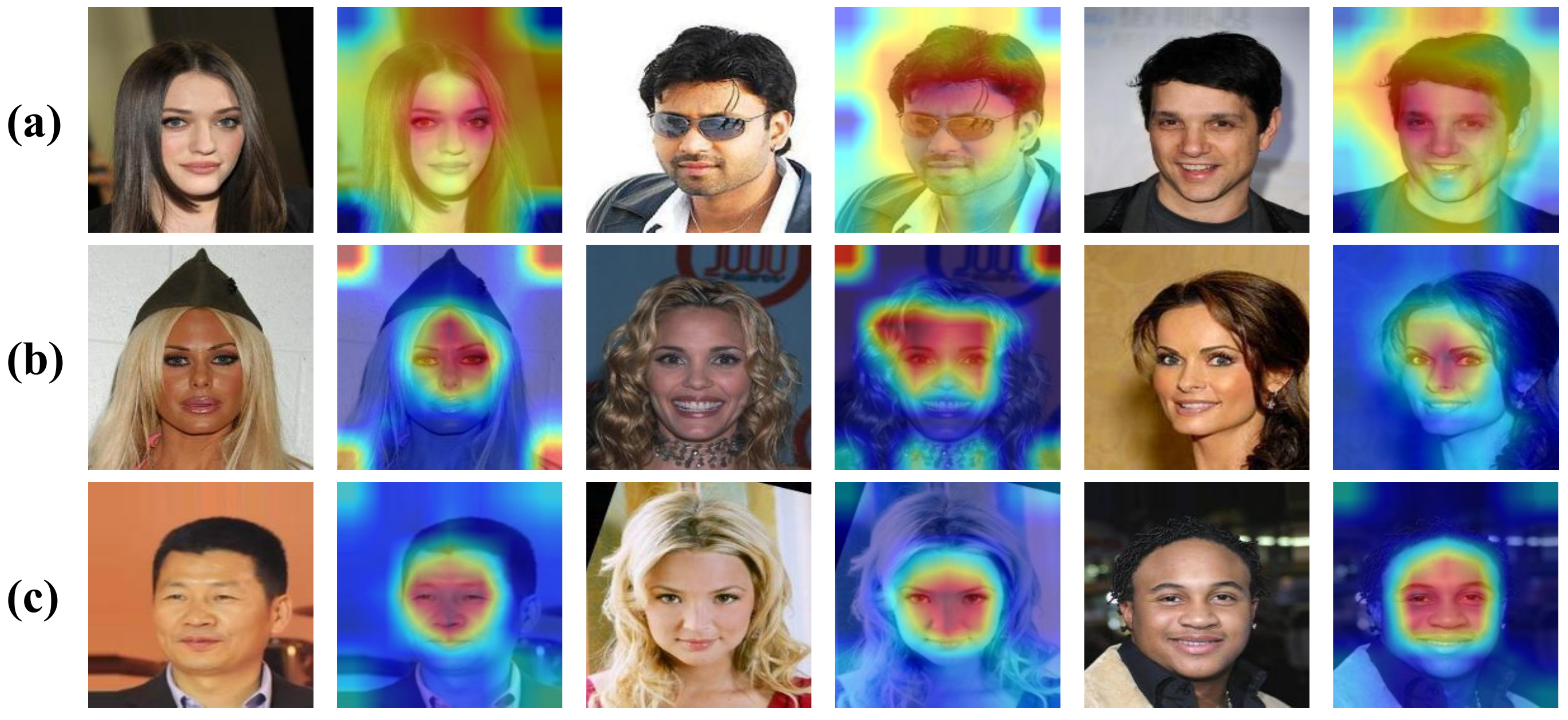}
    \caption{Attention maps and original images of failure cases for (a) Arched eyebrows, (b) Big nose, and (c) 5 o’clock shadow attribute evalution by TransFA on CelebA database. }
    \label{fig:prediction_result}
\end{figure}
As shown in Figure \ref{fig:prediction_result}, we illustrate several failure cases on CelebA database by proposed method. After analyzing the attention maps of these failure cases, we find that TransFA fails to concentrate on the corresponding regions of facial attributes leading to poor evaluation results. For example, Figure \ref{fig:prediction_result} (a) shows that TransFA pays too much attention to the whole face instead of the eyes region when evaluating arched eyebrows. By analyzing the visualization of attention maps,  we could effectively determine whether the evaluation results are reliable.

\section{CONCLUSION} \label{CONCLUSIONS}
The novel transformer-based representation for face attribute evaluation method is proposed in this paper. 
The proposed TransFA could effectively enhance the facial attribute evaluation performance by integrating inter-correlation between different face attributes in similar attention-specific semantic regions. 
Considering integrating the relationship between face identity and semantic attribute information, the hierarchical identity-constraint attribute loss is designed to further improve evaluation performance.
Experimental results on CelebA database and LFWA database demonstrate the superior performance of proposed method compared with SOTA methods. 
The key benefit of our proposed TransFA is  that we utilize a transformer-based representation to capture strong discriminative semantic information for attribute analysis, even for identity recognition.
In the future, we would evaluate the proposed TransFA performance on more large-scale face databases and explore the transformer with convolution in face image analysis.


\bibliographystyle{ACM-Reference-Format}
\bibliography{acmart}

\end{document}